\documentclass[final]{cvpr}

\usepackage{times}
\usepackage{epsfig}
\usepackage{graphicx}
\usepackage{amsmath}
\usepackage{amssymb}
\usepackage{graphicx}
\usepackage{tabularx}
\usepackage{amsthm}
\newcolumntype{Y}{>{\centering\arraybackslash}X}
\usepackage{booktabs}
\usepackage{multirow}
\usepackage{float} 
\usepackage{subfigure}
\usepackage[ruled,linesnumbered]{algorithm2e}

\usepackage[pagebackref=true,breaklinks=true,colorlinks,bookmarks=false]{hyperref}

\def\cvprPaperID{4743} 
\def\confYear{CVPR 2021}

\begin{document}

\title{D$^2$LV: A Data-Driven and Local-Verification Approach for Image Copy Detection}

\author{Wenhao Wang, Yifan Sun, Weipu Zhang, Yi Yang\\
Baidu Research\\
{\tt\small wangwenhao0716@gmail.com, sunyifan01@baidu.com, zhangweipu01@baidu.com, yee.i.yang@gmail.com}
}

\maketitle

\begin{abstract}
\vspace*{-2mm}
Image copy detection is of great importance in real-life social media. In this paper, a data-driven and local-verification ($D^2LV$) approach is proposed to compete for Image Similarity Challenge: Matching Track at NeurIPS’21. In $D^2LV$, unsupervised pre-training substitutes the commonly-used supervised one. When training, we design a set of basic and six advanced transformations, and a simple but effective baseline learns robust representation. During testing, a global-local and local-global matching strategy is proposed. The strategy performs local-verification between reference and query images. Experiments demonstrate that the proposed method is effective. The proposed approach ranks first out of $1,103$ participants on the Facebook AI Image Similarity Challenge: Matching Track. The code and trained models are available \href{https://github.com/WangWenhao0716/ISC-Track1-Submission}{here}.
\vspace*{-4mm}

\end{abstract}

\section{Introduction}

The goal of image copy detection is to determine whether a query image is a modified copy of any images in a reference dataset. It has extensive numbers of applications, such as checking integrity-related problems in social media. Although this topic has been researched for decades, and it has been deemed as a solved problem, most state-of-the-art solutions cannot deliver satisfactory results under real-life scenarios \cite{douze20212021}. There are two main reasons. First, real-life cases in social media involve billions to trillions of images, which introduces many ``distractor” images to degrade the performance. Second, the transformations used to edit images are countless. It is very challenging for an algorithm to be robust to unseen scenarios.\par
\begin{figure}[t]
\centering 
\includegraphics[width=0.47\textwidth]{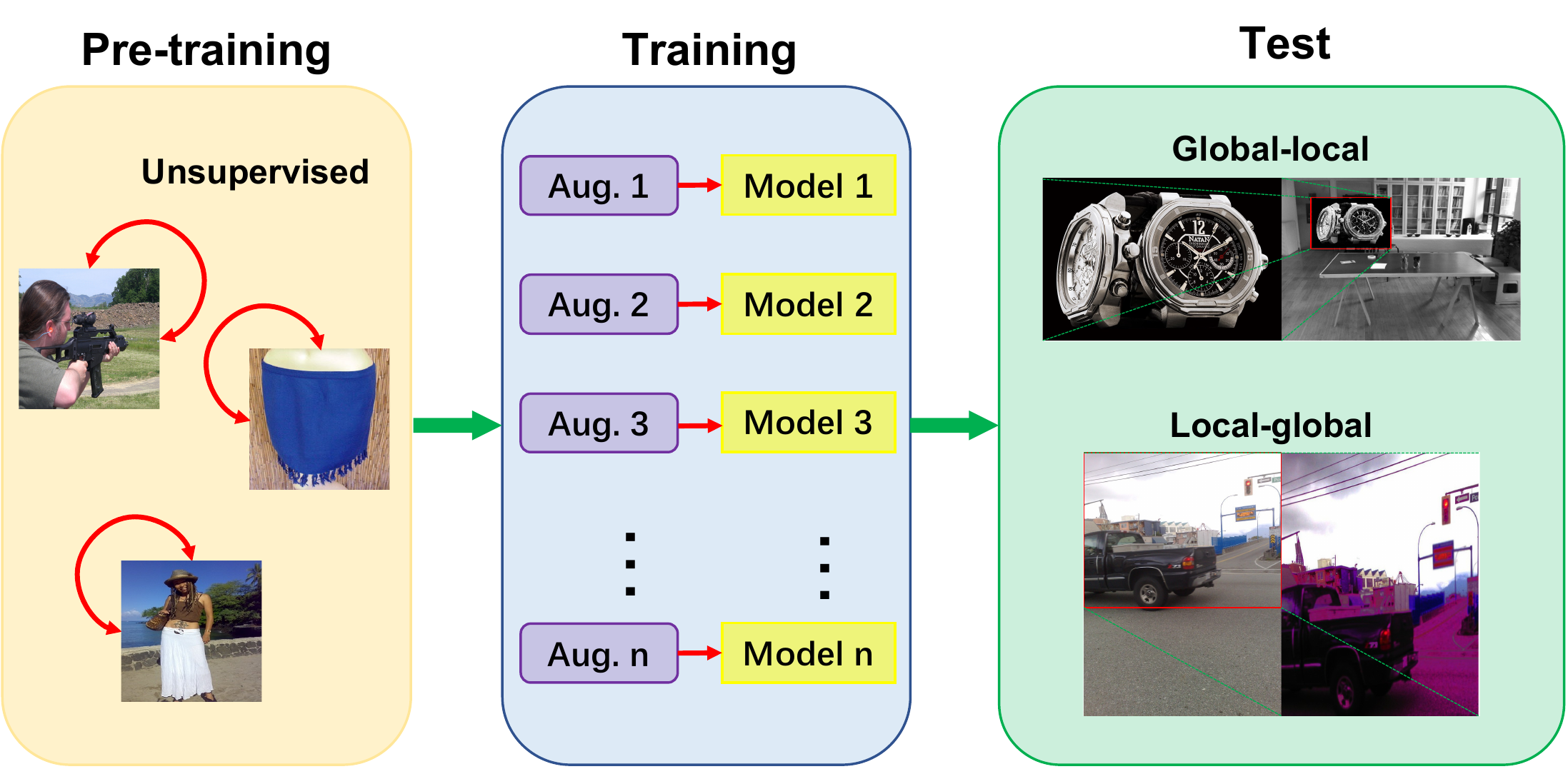}
\vspace*{2mm}
\caption{The data-driven and local-verification approach. Unsupervised pre-training is used to substitute supervised one. During training, we get different trained models by designing different sets of augmentations. Global-local and local-global matching strategy is proposed for testing.} 
\label{frame_1}
\end{figure}
\begin{figure*}[t]
\centering 
\includegraphics[width=0.99\textwidth]{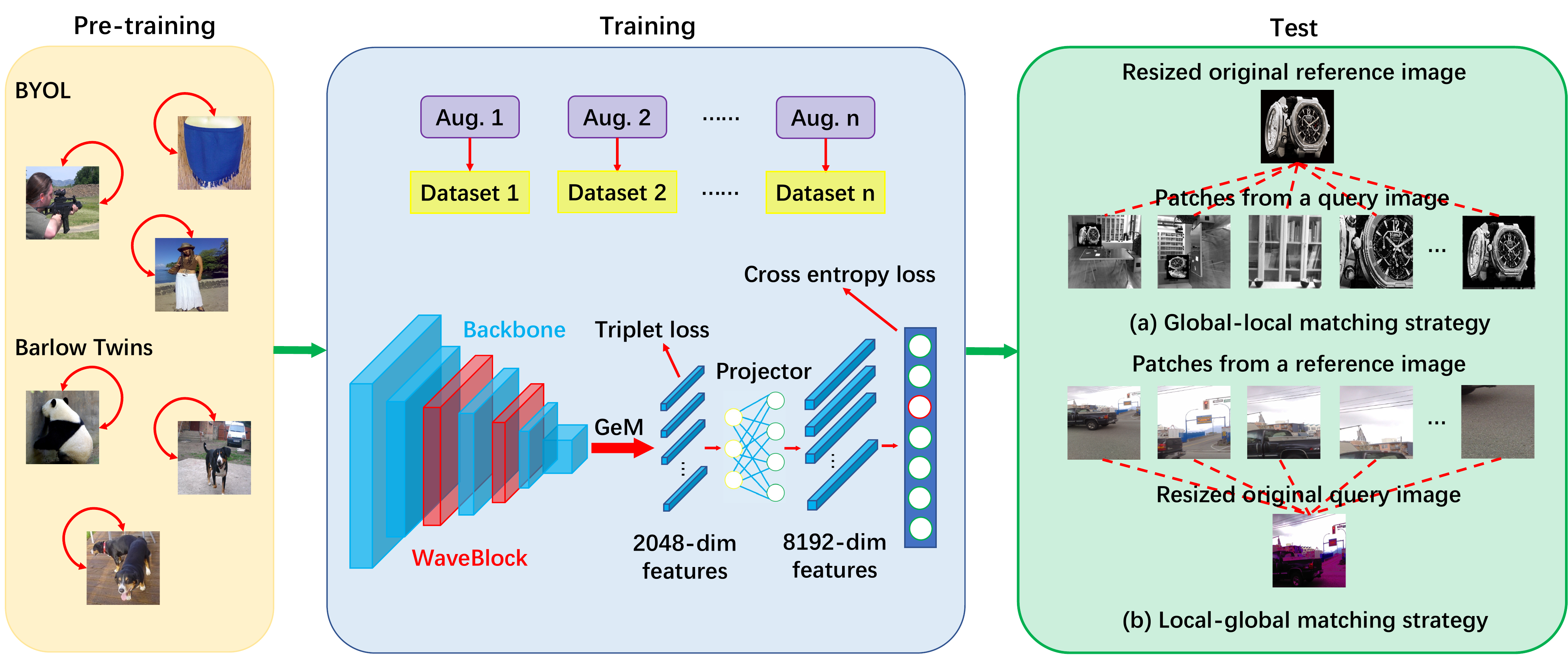}
\vspace*{2mm}
\caption{The proposed data-driven and local-verification (D$^2$LV) approach. Recent self-supervised learning methods, BYOL \cite{jean-bastien2020bootstrap} (its Momentum$^2$ Teacher version \cite{li2021momentum}) and Barlow-Twins \cite{Zbontar2021BarlowTS}, are performed for pre-training. During training, we get different trained models by using the combination of basic augmentations with different advanced augmentations separately. When testing, global-local and local-global matching strategy is proposed for the local-verification between reference and query images.} 
\label{frame_2}
\end{figure*}
In this paper, a data-driven and local-verification (D$^2$LV) approach is proposed to compete for Image Similarity Challenge: Matching Track at NeurIPS'21 (ISC2021) \cite{douze20212021}. This competition builds a benchmark that features a variety of image transformations to mimic real-life cases in social media. To mimic a needle-in-haystack setting, both the query and reference set contain a majority of ``distractor" images that do not match. The evaluation metric adopted is micro Average Precision, which penalizes any detected pair for a distractor query.\par
The approach consists of three parts, \ie pre-training, training, and test. In pre-training, unsupervised pre-training on ImageNet \cite{deng2009imagenet} instead of the commonly-used supervised pre-training is performed. Specifically, we empirically find that BYOL \cite{jean-bastien2020bootstrap} pre-training (its Momentum$^2$ Teacher version \cite{li2021momentum}) and Barlow-Twins \cite{Zbontar2021BarlowTS} pre-training are superior to some other unsupervised pre-training methods, and a further ensemble of these two is even better. In training, a strong but simple deep metric learning baseline is designed by combining both the classification loss and the pairwise loss (triplet, in particular). Moreover, to make the sample pairs more informative, a battery of image augmentations is employed to generate training images. Empirically, we find that different augmentations may disturb each other to degrade the performance. Therefore, we use different sets of augmentations to train models separately and then ensemble them. The diversity of augmentations promotes learning robust representation. During testing, a robust global-local and local-global matching strategy is proposed. We observe two hard cases: a) some query images are generated by overlaying a reference image on top of a distractor image and b) some queries are cropped from the reference and thus contain only a partial patch of the reference images. In response, both heuristic and auto-detected bounding boxes are adopted to crop some local patches for global-local and local-global matching. The illustration of the proposed approach is shown in Fig. \ref{frame_1}. \par

In summary, the main contributions of this paper are:
\begin{enumerate}
 \item The paper proposes a data-driven and local-verification approach for image copy detection.
 \item The proposed D$^2$LV approach handles real-life copy detection scenarios in social media well and generalizes well to unseen transformations.
 \item Our approach outperforms the baseline models and achieves the state-of-the-art in the Image Similarity Challenge: Matching Track at NeurIPS’21.
\end{enumerate}

\section{Related Work}

\subsection{Copy Detection}
Although copy detection plays an important role in social media, the publication of copy detection is not well known because organizations want to keep copy detection techniques obscure, and the researchers often consider that the task is easy \cite{douze20212021}. For classical approaches, some methods are used to extract global \cite{kim2003content, wan2008survey, hsiao2007new} and local descriptors \cite{amsaleg2001content, berrani2003robust, ke2004efficient}. For deep learning methods, a descriptor, or known as a feature, is extracted by convolutional neural networks \cite{liu2020content, zhou2020cnn}. However, none of them gives a satisfying performance on challenging, large benchmarks like ISC2021.
\subsection{Unsupervised Pre-training}
The unsupervised pre-trained models are from recent self-supervised learning methods. They are trained to extract image embedding unsupervisedly. MoCo \cite{he2020momentum} builds a dynamic dictionary with a queue and a moving-averaged encoder to conduct contrastive learning. BYOL \cite{jean-bastien2020bootstrap} and its Momentum$^2$ Teacher version  achieve a new state-of-the-art without negative pairs. Other self-supervised methods, such as Barlow Twins \cite{Zbontar2021BarlowTS}, SimSiam \cite{chen2021exploring}, and SwAV \cite{caron2020unsupervised}, also show promising performance. \par 
For copy detection in ISC2021, we find that unsupervised pre-trained models show superior performance to their fully-supervised counterparts. That may be because the category defined in unsupervised pre-training is much more similar to that in copy detection than in fully-supervised one.
\begin{figure*}[t]
\centering 
\includegraphics[width=1\textwidth]{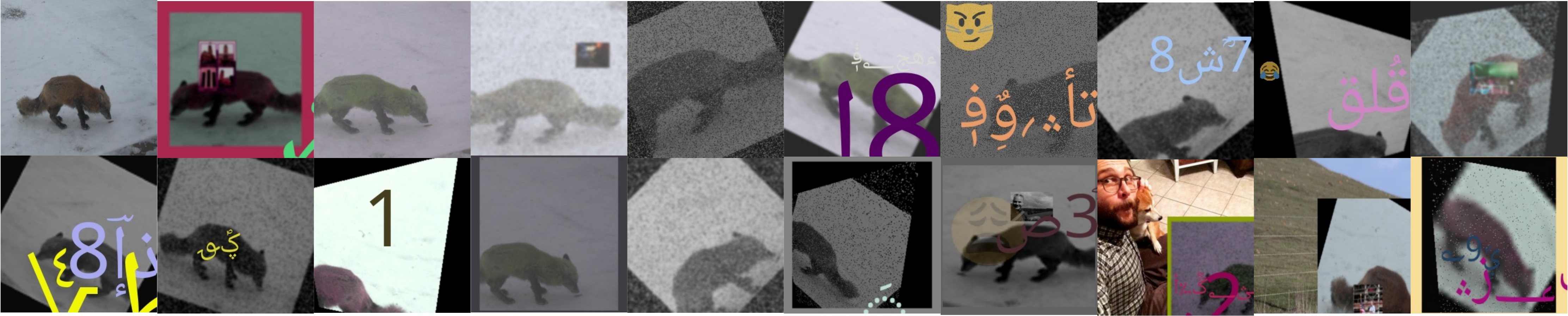}
\vspace*{2mm}
\caption{The set of basic augmentations. It includes random resized cropping, random rotation, random pixelization, random pixels shuffling, random perspective transformation, random padding, random image underlay, random color jitter, random blurring, random gray scale, random horizontal flipping, random Emoji overlay, random text overlay, random image overlay, and resizing. The first image is the resized original image.} 
\label{basic}
\end{figure*}
\subsection{Deep Metric Learning}
Learning a discriminative feature is a crucial component for many tasks, such as image retrieval \cite{qian2019softtriple, kim2020proxy}, face verification \cite{Sun_2020_CVPR, wang2017normface}, and object re-identification \cite{sun2018beyond, zheng2020vehiclenet}. The commonly-used loss functions can be divided into two classes: pair-based \cite{hermans2017defense, sohn2016improved, oh2016deep} and proxy-based \cite{Liu_2017_CVPR, wang2018cosface, deng2019arcface} losses. Circle loss \cite{Sun_2020_CVPR} gives a unified formula for two paradigms. \par 
In our solution to ISC2021, we only use two common losses, \ie triplet loss with hard sample mining and cross-entropy loss with label smoothing \cite{szegedy2016rethinking}. The two losses are proved to be simple but effective.
\section{Proposed Method}
In this section, we introduce each important component in the proposed D$^2$LV approach. In the training part, we discuss the built simple but effective baseline and designed different sets of augmentations. In the test part, we discuss the local verification and ensemble methods. The whole approach is shown in Fig. \ref{frame_2}.

\subsection{Unsupervised Pre-training}
In the ISC2021, a category is defined at a very ``tight" level, \ie images generated by exact duplication, near-exact duplication, and edited copy are considered in the same category while two images that belong to the same instance or object are not in the same category. However, when using fully-supervised pre-training on the ImageNet, two images share the same instance or object are in one class. Therefore, the definition of the category is contradicted between the ISC2021 and ImageNet. \par
Fortunately, the recent research on self-supervised learning provides a new direction. It defines every image as a category and uses invariance to data augmentation as their main training criterion \cite{douze20212021}. Although it seems natural to directly adopt a self-supervised learning method to train a model for copy detection, we choose to give up this solution for two reasons. First, the self-supervised learning methods often cost a lot of days to get convergence results. For instance, on ImageNet, BYOL \cite{jean-bastien2020bootstrap} or its Momentum$^2$ Teacher version \cite{li2021momentum} takes about two weeks to pre-train $300$ epochs with ResNet-152 \cite{he2016deep} backbone using $8$ NVIDIA Tesla V100 32GB GPUs. In addition, when using sophisticated transformations online, the time becomes much longer, which is unaffordable. Second, even if the resource is unlimited, we do not get a satisfying performance only trained by  self-supervised learning methods, which may be because that the hyper-parameters are not adjusted carefully.\par
As a result, we choose to use BYOL \cite{jean-bastien2020bootstrap} (its Momentum$^2$ Teacher version \cite{li2021momentum}) and Barlow-Twins \cite{Zbontar2021BarlowTS} to get the pre-trained model on ImageNet. The used augmentations follow the default setting in their original implementations.

\subsection{Baseline}
We design a simple but effective baseline for ISC2021. Its original version is from DomainMix \cite{wang2021domainmix}. The built baseline newly includes GeM \cite{radenovic2018fine}, WaveBlock \cite{wang2020attentive}, a high-dimension projector, two commonly-used losses, and a warming up with cosine annealing learning rate. The backbones selected are ResNet-50 \cite{he2016deep}, ResNet-152 \cite{he2016deep}, and ResNet50-IBN \cite{Pan_2018_ECCV}. For the design of GeM \cite{radenovic2018fine} and WaveBlock \cite{wang2020attentive}, please refer to their original papers, and we follow their default hyper-parameters. The high-dimension projector projects a learned $2048$-dim feature into $8192$-dim via some linear and non-linear connections. We empirically find learning from high dimension space is beneficial for datasets with large-scale categories. The two commonly-used losses are triplet loss with hard sample mining and cross-entropy loss with label smoothing \cite{szegedy2016rethinking}. The $2048$-dim feature is used for triplet loss, and the $8192$-dim feature is used for classification. The baseline is trained for $25$ epochs, and the change of ratio for learning rate is:
\begin{equation}
ratio = \left\{ \begin{array}{l}
0.99 \cdot epoch/5 + 0.01,0 \le epoch < 5\\
1,5 \le epoch < 10\\
0.5 \cdot \left( {\cos \left( {{\textstyle{{epoch - 10} \over {25 - 10}}} \cdot \pi } \right) + 1} \right),10 \le epoch < 25
\end{array} \right.
\end{equation}

\begin{figure}

\centering

\subfigure[Super-blur augmentation.]{

\begin{minipage}[t]{0.47\textwidth}

\includegraphics[width=1\textwidth]{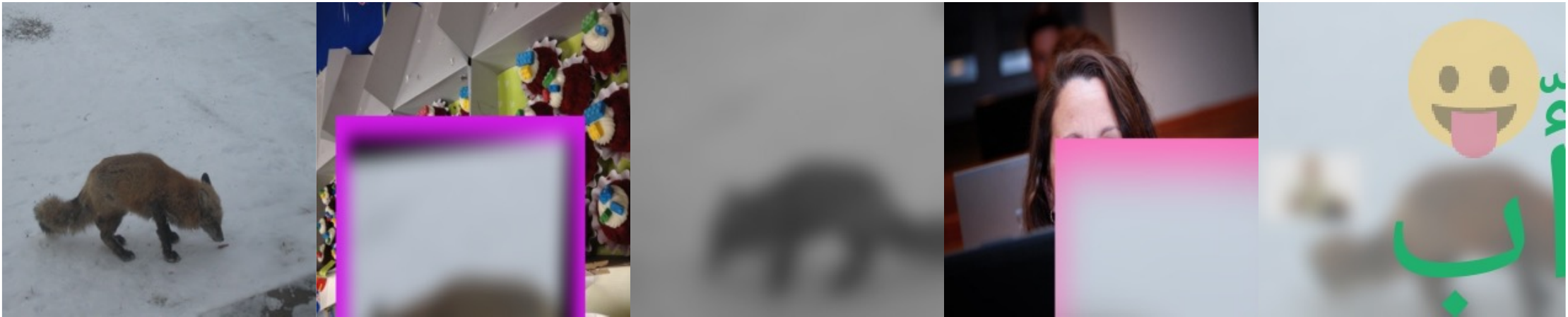}

\end{minipage}

}

\subfigure[Super-color augmentation.]{

\begin{minipage}[t]{0.47\textwidth}

\includegraphics[width=1\textwidth]{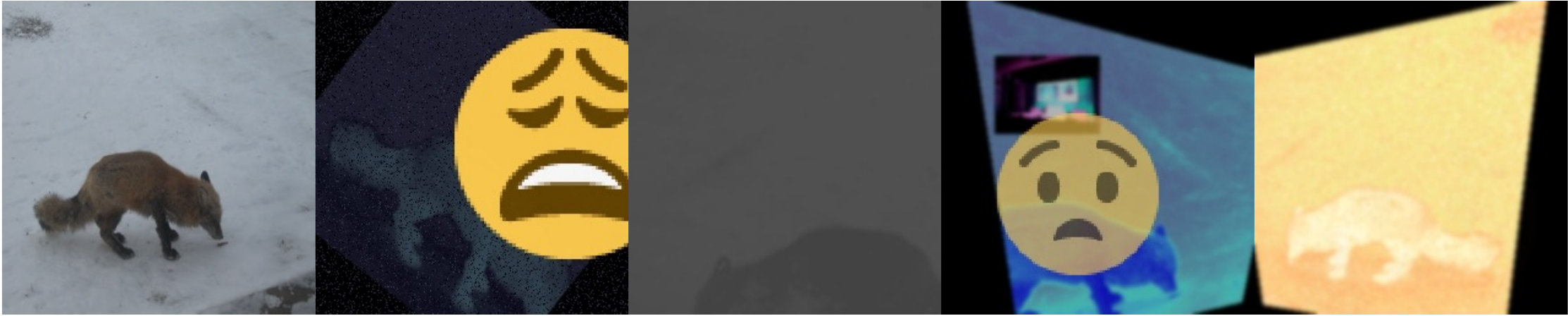}

\end{minipage}

}

\subfigure[Super-dark augmentation.]{

\begin{minipage}[t]{0.47\textwidth}

\includegraphics[width=1\textwidth]{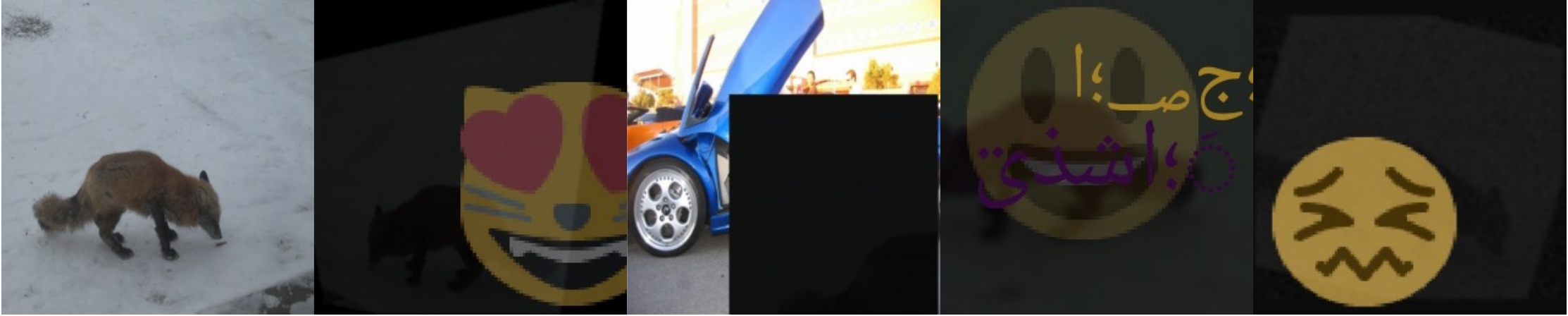}

\end{minipage}

}

\subfigure[Super-face augmentation.]{

\begin{minipage}[t]{0.47\textwidth}

\includegraphics[width=1\textwidth]{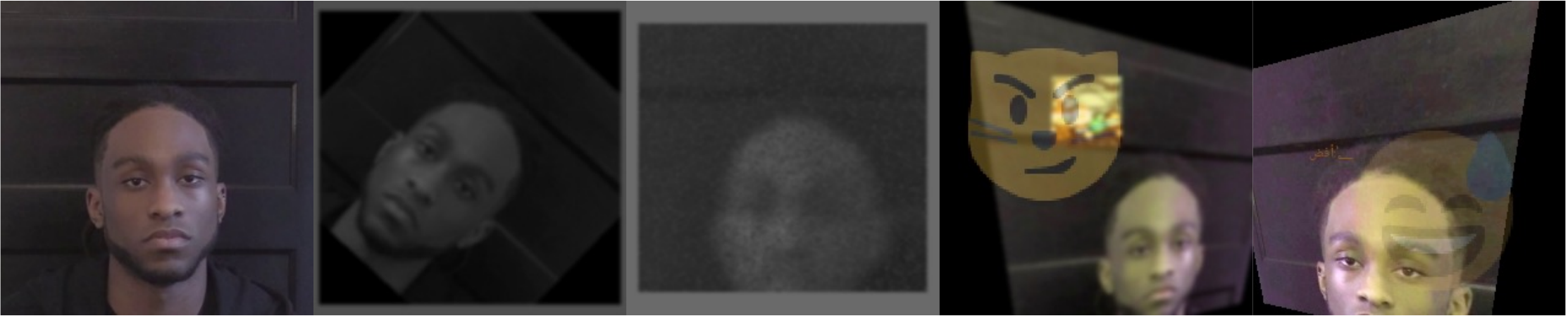}

\end{minipage}

}

\subfigure[Super-opaque augmentation.]{

\begin{minipage}[t]{0.47\textwidth}

\includegraphics[width=1\textwidth]{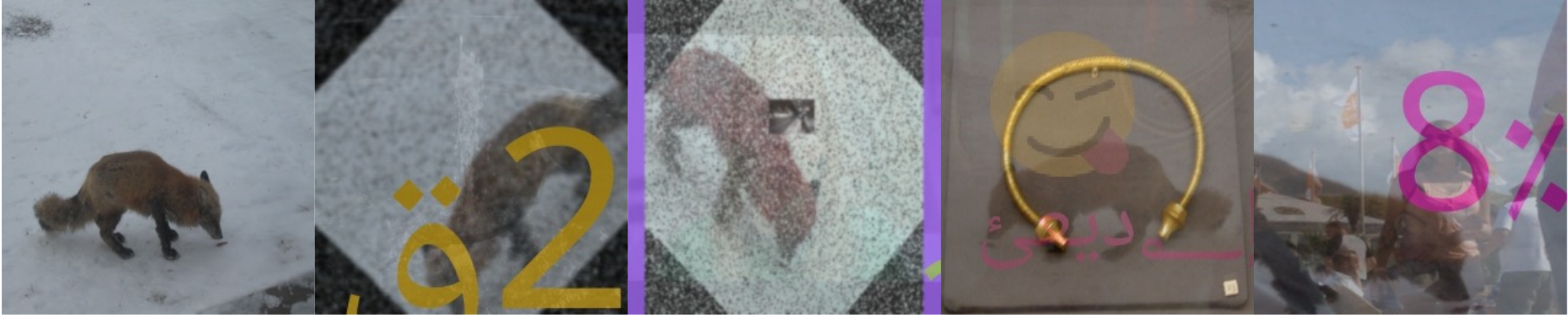}

\end{minipage}

}

\subfigure[Super-occlude augmentation.]{

\begin{minipage}[t]{0.47\textwidth}

\includegraphics[width=1\textwidth]{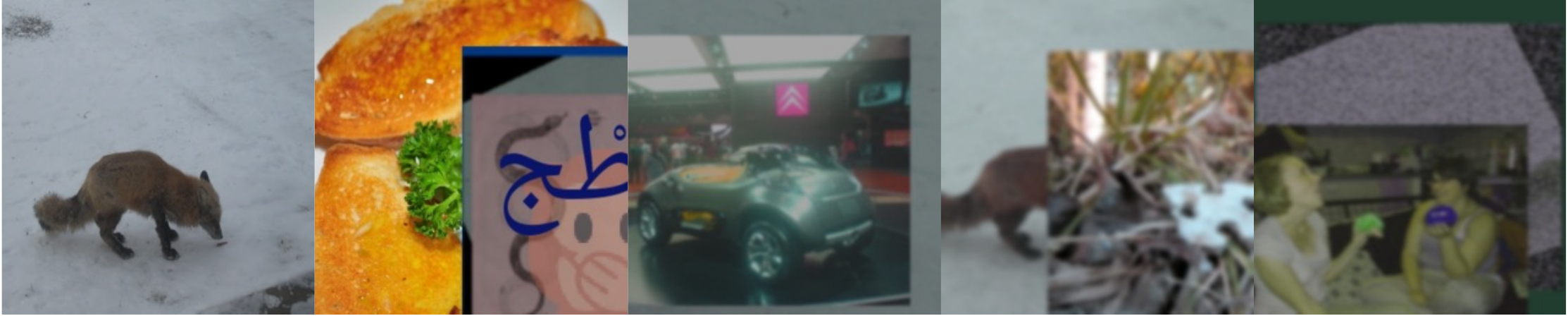}

\end{minipage}

}

 \caption{The six advanced augmentations. They are added into the original set of basic augmentations separately. The first image is the resized original image.} \label{six_advanced}

\end{figure}
\subsection{Augmentation}
In the ISC2021, the data is crucial in our solution. That is why the proposed approach is named ``data-driven''. First, we design a set of basic augmentations to transform images. Then, we build six advanced augmentations to provide a supplementary.\par 
The set of basic augmentations includes random resized cropping, random rotation, random pixelization, random pixels shuffling, random perspective transformation, random padding, random image underlay, random color jitter, random blurring, random grayscale, random horizontal flipping, random Emoji overlay, random text overlay, random image overlay, and resizing. A display of using all the basic augmentations on one image is shown in Fig. \ref{basic}. The first image is the resized original image. \par
Further, together with the set of basic transformations, we design six advanced augmentations, \ie super-blur, super-color, super-dark, super-face, super-opaque, and super-occlude. The six augmentations are added into the set of basic augmentations \textbf{\textit{separately}}. The super-blur augmentation uses enhanced blur augmentation; the super-color augmentation uses enhanced color jitter augmentation; the super-dark augmentation darkens the images; the super-face augmentation adds some face images into training; the super-opaque augmentation overlays one image on another with certain transparency; the super-occlude augmentation adds more occlusion into original images. These six advanced augmentations are shown in Fig. \ref{six_advanced}. \par
Besides the basic and advanced augmentations, we also employ black-white augmentation. The augmentation changes all the color images into black and white style. Some of the sets, \ie ``basic'', ``basic + super-blur'', ``basic + super-color'', and ``basic + super-face'', use this augmentation. As a result, we have $11$ sets containing different augmentations. A pre-trained model is trained $11$ times to get $11$ trained models.

\subsection{Local Verification}
During testing, we observe two corner cases: a) some query images are generated by overlaying a reference image on top of a distractor image. b) some queries are cropped from the reference images and thus only contain parts of the reference images. Therefore, we propose a global-local and local-global matching strategy for testing. For global-local matching, it matches the global features of reference images with local features of query images. For local-global matching, it matches the local features of reference images with global features of query images.\par
\begin{figure}[t]
\centering 
\includegraphics[width=0.47\textwidth]{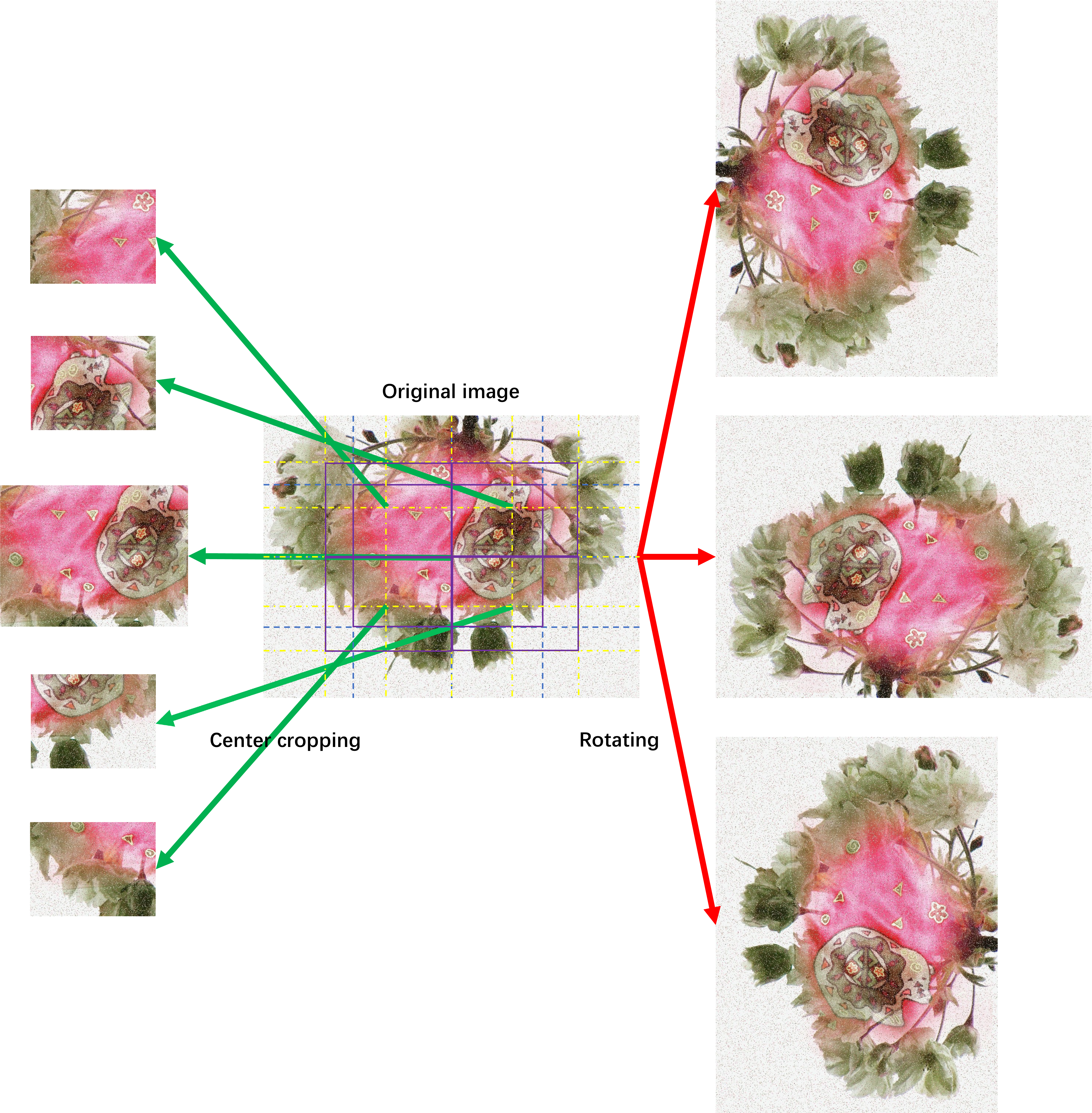}
\vspace*{2mm}
\caption{The nine patches generated by rotating and center cropping. The dotted lines in the original image show the definition of ``exact center" and ``$1/3$ center''.} 
\label{patch}
\end{figure}

 
\textbf{Generating local features of query images.} We use heuristic and auto-detected bounding boxes to crop local patches from query images. The features of cropped local patches are used as local features of query images. Given a query image, first, rotating and center cropping are used to generate patches. The rotating degrees are $90^{\circ}$, $180^{\circ}$, and $270^{\circ}$. The center cropping crops not only the ``exact center" but also the ``$1/3$ center'' of an image. The illustration of ``exact center" and ``$1/3$ center'' is shown in Fig. \ref{patch}. Together with the original image, now nine patches are generated. Second, selective search \cite{uijlings2013selective} is used to generate proposal regions. The features of proposal regions are also regarded as local features of query images. Last, we use YOLOv5 \cite{glenn_jocher_2020_4154370} to detect overlay automatically. For training, we just generate some images with overlay augmentations and corresponding bounding boxes automatically from the training dataset. Note that the above-defined “patch” includes the original image and its rotations.\par 
\textbf{Generating local features of reference images.} We only divide reference images into small patches to match locally. First, given a reference image, it is divided into four parts evenly and nine parts evenly. Then, we crop the ``exact center" and ``$1/3$ center'' of a reference image. Together with the original image, we can extract $19$ features from a reference image. Note that the above-defined ``patch'' also includes the original image.

\subsection{Ensemble}
First, we introduce two ensemble criteria, \ie confidence and completeness. Assume that we have a pair of images, e.g. a patch $A$ from a query image and an original reference image $B$, and two models, $f$ and $g$. $f\left(A,B\right)$ and $g\left(A,B\right)$ stand for the similarity score between $A$ and $B$ using models $f$ and $g$, respectively. Also, two score thresholds are $\alpha$ and $\beta$. When using confidence criterion, the ensemble result is:

\begin{footnotesize}
\begin{equation}
score=\begin{cases}max(f\left( A,B\right)  ,g\left( A,B\right)  ),\  if\  f\left( A,B\right)  >\alpha \  and\  g\left( A,B\right)  >\beta ,&\  \\ None,\  others.&\end{cases} \  
\end{equation}
\end{footnotesize}

The ``None'' means both the scores are discarded, \ie it does not contribute to the final score of the image pair.
Under completeness criterion, the ensemble result is:
\begin{equation}
score=max(f\left( A,B\right)  ,g\left( A,B\right)  ).
\end{equation}
Both the confidence and completeness criteria can be extended to multi-models by

\begin{footnotesize}
\begin{equation}
score=\begin{cases}max(f_{0}\left( A,B\right)  ,f_{1}\left( A,B\right)  ,...,f_{n}\left( A,B\right)  ),\  if\  f_{i}\left( A,B\right)  >\alpha_{i} ,&\  \\ None,\  others,&\end{cases} 
\end{equation}
\end{footnotesize}

and
\begin{equation}
score=max(f_{0}\left( A,B\right)  ,f_{1}\left( A,B\right)  ,...,f_{n}\left( A,B\right) ),
\end{equation}
where: $i$ is from $0$ to $n$, $f_{0},f_{1},...,f_{n}$ denotes $n$ different models, and $\alpha_{0} ,\alpha_{1} ,...,\alpha_{n} $ denotes $n$ different score thresholds. \par
Remember that we have three different backbones, when using the global-local matching strategy, the confidence criterion is used for ResNet50$\&$ResNet152, ResNet50$\&$ResNet50-IBN, and ResNet152$\&$ResNet50-IBN. When using the local-global matching strategy, the confidence criterion is used for ResNet50$\&$ResNet152$\&$ResNet50-IBN. For any other model ensembles, we use the completeness criterion. \par
Further, to ensemble different local-global and global-local scores, we repeat the completeness criterion by 
\begin{equation}
score=max\left( \begin{gathered}max(f\left( A_{0},B\right)  ,\  f\left( A_{1},B\right)  ,...,f\left( A_{l},B\right)  ),\\ max(f\left( A,B_{0}\right)  ,\  f\left( A,B_{1}\right)  ,...,f\left( A,B_{m}\right)  )\end{gathered} \right)  
\end{equation}
where: $A_{0},A_{1},...,A_{l}$ denotes $l$ different patches of the query image, and $B_{0},B_{1},...,B_{m}$ denotes $m$ different patches of the reference image. \par 
After finishing all models and patches ensemble, the final score represents the similarity between the pair of images.

\section{Experiments}

\subsection{Experimental Settings}
In the unsupervised pre-training part, we essentially follow the experimental setting given in Momentum$^2$ Teacher \cite{li2021momentum} and Barlow-Twins \cite{Zbontar2021BarlowTS}. For Momentum$^2$ Teacher \cite{li2021momentum}, the selected backbone is ResNet152 \cite{he2016deep} and ResNet50-IBN \cite{Pan_2018_ECCV}. $8$ NVIDIA Tesla V100 32GB GPUs are used. When training ResNet152 \cite{he2016deep}, the batch size is $512$, and it is trained for $300$ epochs. The training process can be finished in about two weeks. When training ResNet50-IBN \cite{Pan_2018_ECCV}, the batch size is $1024$, and it is also trained for $300$ epochs. The training process can be finished in about six days. For Barlow-Twins \cite{Zbontar2021BarlowTS}, the selected backbone is ResNet50 \cite{he2016deep}. We do not re-train the model, and just use the model supplied by the official implement. \par
In training, we select $100,000$ out of $1000,000$ images from the official training set. The criterion is selecting one every $10$. Each image is augmented $19$ times, and thus for training, each ID has $20$ images. We trained $33$ models in total. Each training takes about/less than one day on $4$ NVIDIA Tesla V100 32GB GPUs (bigger backbone takes longer and smaller backbone takes shorter). Each training batch includes $128$ images of $32$ IDs. Adam optimizer \cite{DBLP:journals/corr/KingmaB14} is used to optimize the networks. The image size is $256 \times 256$. The model is trained for $25$ epochs, and in each epoch, we have $8000$ iterations. The basic learning rate is set to $3.5 \times 10^{-4}$.\par
During testing, for global-local matching, we test the $33$ trained models for three scales, \ie $200 \times 200$, $256 \times 256$, $320 \times 320$, then ensemble them to get the final results. The time for extracting all query patches' features using $33$ NVIDIA Tesla V100 32GB GPUs is about six hours. For local-global matching, we only test the three models trained on the basic augmentations for one scale, \ie $256 \times 256$. The time for extracting all $19,000,000$ reference patches’ features using $3$ NVIDIA Tesla A100 40GB GPUs on a DGX-server is less than one day. We apply a PCA dimensionality reduction to decrease dimension from $8192$ to $1500$ with the code from the competition official. The $1500$-dim feature is used for final matching. Besides, some tricks are used to further improve performance. First, when performing local-global strategy, it is inappropriate to divide reference images containing faces into patches. Therefore, when a face detector finds face(s) existing in one reference image, we do not perform the local-global strategy. Second, compared to original images, the partial image in a pair is more likely to be ``false positive'', and thus we penalize the corresponding score. Third, we delete generated query patches with too small image size. Fourth, the final ensemble score uses the average of two maximum scores instead of the only one as introduced before.
\subsection{Comparison with State-of-the-Arts}
\begin{table}
  \caption{Comparison with state-of-the-art methods from the leaderboard in the Phase $2$. Recall@Precision $90$ is a secondary metric provided for information purposes only as an indicator of a model's recall at a reasonably good precision level, but is not used for ranking purposes. Our results are highlighted in bold.}
  \vspace*{2mm}
\small
  \begin{tabularx}{\hsize}{|p{1.5cm}|YY|}
    \hline
    \multicolumn{1}{|c|}{\multirow{2}{*}{Team}} &
    \multicolumn{2}{c|}{Score}  \\
    \cline{2-3}
      &\footnotesize{Micro-average Precision} & \footnotesize{Recall@Precision 90} \\
    \hline\hline
    \textbf{Ours} & $\textbf{0.8329}$ & $\textbf{0.7309}$  \\
    separate& $0.8291$ & $0.7917$   \\
    imgFp  & $0.7682$ & $0.6715$ \\
    forthedream  & $0.7667$ & $0.7218$ \\
    titanshield  & $0.7613$ & $0.7557$ \\
    VisonGroup  & $0.7169$ & $0.5963$ \\
    mmcf & $0.7107$ & $0.5986$ \\
    ...&...&... \\
    \hline
    \footnotesize{MultiGrain\cite{2019arXivMultiGrain}} & $0.2761$ & $0.2023$ \\
    GIST \cite{oliva2001modeling} & $0.0526	$ & $-$ \\ 
    \hline
  \end{tabularx}

  \label{sota}
\end{table}
To prove the superiority of the D$^2$LV, we compare the proposed model with state-of-the-art methods from the leaderboard in Phase $2$. The comparison results are shown in Table \ref{sota}. In the ISC2021, there are $1103$ participants, and $36$ teams have submitted their final results. Compared with the strongest official benchmark MultiGrain \cite{2019arXivMultiGrain}, the proposed D$^2$LV achieves about $56\%$ improvements, which is considerable. Most Micro-average Precisions from other teams are less than $77\%$, while our approach gives an $83.29\%$ result. It is interesting to mention that, in Phase $1$, our result is $90.08\%$. There is only a less than $7\%$ gap between the two leaderboards for us. However, for some other teams, the gap can be larger than $30\%$. The superiority performance is attributed to two aspects. First, the sophisticated augmentations give us more generalizability when facing unseen transformations. That is called ``Data-Driven''. Second, the global-local and local-global matching strategy provides an exhaustive matching between a pair of images. That is called ``Local-Verification''.

\subsection{Ablation Studies}
The ablation studies are conducted on $25,000$ query images in Phase $1$ to prove the effectiveness of each component in the D$^2$LV approach. The evaluation metric is Micro-average Precision. The experimental results are displayed in Table \ref{abla}. \par

 \begin{table}
\small
\caption{The ablation study about the proposed D$^2$LV. ``Supervised" or ``Unsupervised" denotes the supervised pre-training or unsupervised pre-training is used, respectively. ``Global-local'' denotes using global-local matching strategy, and ``Both' denotes using global-local and local-global matching strategy. ``Adv-Aug'' denotes using different kinds of advanced augmentations. ``Multi+Tricks'' denotes using multi-scale testing and other mentioned tricks. The best results are highlighted in bold.}
\vspace*{2mm}
  \begin{tabularx}{\hsize}{|p{1.5cm}|YY|}
    \hline
    \multicolumn{1}{|c|}{\multirow{2}{*}{Method}} &
    \multicolumn{2}{c|}{Score}  \\
    \cline{2-3}
      &\footnotesize{Micro-average Precision} & \footnotesize{Recall@Precision 90} \\
    \hline\hline
    Supervised & $0.68726$ & $0.54678$  \\
    Unsupervised& $0.70813$ & $0.62773$   \\
    \hline
    \footnotesize{Global-local}& $0.82726$ & $0.74755$ \\
    Both  & $0.83720$ & $0.75155$ \\
    \hline
    Adv-Aug & $0.88640$ & $0.80124$ \\
    \hline
    Multi+Tricks & $\textbf{0.90035}$ & $\textbf{0.81887}$ \\
    \hline
  \end{tabularx}

  \label{abla}
\end{table}
\textbf{Comparison between supervised pre-training and unsupervised pre-training.}
First, we discuss the improvement brought by unsupervised pre-training. The experimental results are denoted as ``Supervised" and ``Unsupervised'' in Table \ref{abla}, respectively. The training images only use the basic augmentations. The matching strategy is only global-to-global. We can find that, instead of supervised pre-training, $2.1\%$ improvement has been obtained by using unsupervised pre-trained models. This phenomenon may be because the definition of category in unsupervised pre-training is similar to that in the task of copy detection. \par 
\textbf{The improvement from local-verification.} In this part, the experimental results are from the ensemble result of the three backbones. The experimental results are denoted as ``Global-local" and ``Both'' in Table \ref{abla}, respectively. When only using the global-local matching strategy, the performance can be improved from $70.81\%$ to $82.73\%$. When we perform global-local and local-global matching strategy, the highest performance, \ie $83.72\%$, is achieved. The local-verification plays a very important role in the proposed D$^2$LV. \par 
\textbf{The improvement from advanced augmentations.} To be convenient, we only give the final ensemble experimental results of all the advanced augmentations with the basic ones. The experimental results are denoted as ``Adv-Aug'' in Table \ref{abla}. Indeed, each advanced augmentation with the basic one can improve performance separately. Also, the results use local-verification and three backbones. The sophisticated augmentations contribute a lot to Phase 2 where some unseen transformations appear.\par 
\textbf{The improvement from multi-scale testing and some tricks.}
The experimental results are denoted as ``Multi+Tricks'' in Table \ref{abla}. The performance improved by multi-scale testing and some tricks is not very significant. It is less than $1.5\%$ on $25,000$ queries in Phase 1. After finding the huge performance gap between the two phases, we are not quite sure whether these methods really contribute to the final results in Phase 2.

\section{Conclusion}
In this paper, we introduce our winning solution to Image Similarity Challenge: Matching Track at NeurIPS'21. The proposed D$^2$LV approach uses recent self-supervised learning methods for pre-training instead of traditional supervised ones. Further, we find that exploring data is of great importance, and thus many strong augmentations are designed. Also, the proposed global-local and local-global matching strategy contributes a lot to the final submission. We hope the proposed solution is beneficial for real-life applications including content tracing, copyright infringement, and misinformation.

{\small
\bibliographystyle{ieee_fullname}
\bibliography{egbib}

\begin{thebibliography}{10}\itemsep=-1pt

\bibitem{amsaleg2001content}
Laurent Amsaleg and Patrick Gros.
\newblock Content-based retrieval using local descriptors: Problems and issues
  from a database perspective.
\newblock {\em Pattern Analysis \& Applications}, 4(2):108--124, 2001.

\bibitem{2019arXivMultiGrain}
Maxim Berman, Herv{\'e} J{\'e}gou, Vedaldi Andrea, Iasonas Kokkinos, and
  Matthijs Douze.
\newblock {{MultiGrain}: a unified image embedding for classes and instances}.
\newblock {\em arXiv e-prints}, Feb 2019.

\bibitem{berrani2003robust}
Sid-Ahmed Berrani, Laurent Amsaleg, and Patrick Gros.
\newblock Robust content-based image searches for copyright protection.
\newblock In {\em Proceedings of the 1st ACM international workshop on
  Multimedia databases}, pages 70--77, 2003.

\bibitem{caron2020unsupervised}
Mathilde Caron, Ishan Misra, Julien Mairal, Priya Goyal, Piotr Bojanowski, and
  Armand Joulin.
\newblock Unsupervised learning of visual features by contrasting cluster
  assignments.
\newblock 2020.

\bibitem{chen2021exploring}
Xinlei Chen and Kaiming He.
\newblock Exploring simple siamese representation learning.
\newblock In {\em Proceedings of the IEEE/CVF Conference on Computer Vision and
  Pattern Recognition}, pages 15750--15758, 2021.

\bibitem{deng2009imagenet}
Jia Deng, Wei Dong, Richard Socher, Li-Jia Li, Kai Li, and Li Fei-Fei.
\newblock Imagenet: A large-scale hierarchical image database.
\newblock In {\em 2009 IEEE conference on computer vision and pattern
  recognition}, pages 248--255. Ieee, 2009.

\bibitem{deng2019arcface}
Jiankang Deng, Jia Guo, Niannan Xue, and Stefanos Zafeiriou.
\newblock Arcface: Additive angular margin loss for deep face recognition.
\newblock In {\em Proceedings of the IEEE/CVF Conference on Computer Vision and
  Pattern Recognition}, pages 4690--4699, 2019.

\bibitem{douze20212021}
Matthijs Douze, Giorgos Tolias, Ed Pizzi, Zo{\"e} Papakipos, Lowik Chanussot,
  Filip Radenovic, Tomas Jenicek, Maxim Maximov, Laura Leal-Taix{\'e}, Ismail
  Elezi, et~al.
\newblock The 2021 image similarity dataset and challenge.
\newblock {\em arXiv preprint arXiv:2106.09672}, 2021.

\bibitem{he2020momentum}
Kaiming He, Haoqi Fan, Yuxin Wu, Saining Xie, and Ross Girshick.
\newblock Momentum contrast for unsupervised visual representation learning.
\newblock In {\em Proceedings of the IEEE/CVF Conference on Computer Vision and
  Pattern Recognition}, pages 9729--9738, 2020.

\bibitem{he2016deep}
Kaiming He, Xiangyu Zhang, Shaoqing Ren, and Jian Sun.
\newblock Deep residual learning for image recognition.
\newblock In {\em Proceedings of the IEEE conference on computer vision and
  pattern recognition}, pages 770--778, 2016.

\bibitem{hermans2017defense}
Alexander Hermans, Lucas Beyer, and Bastian Leibe.
\newblock In defense of the triplet loss for person re-identification.
\newblock {\em arXiv preprint arXiv:1703.07737}, 2017.

\bibitem{hsiao2007new}
Jen-Hao Hsiao, Chu-Song Chen, Lee-Feng Chien, and Ming-Syan Chen.
\newblock A new approach to image copy detection based on extended feature
  sets.
\newblock {\em IEEE Transactions on Image Processing}, 16(8):2069--2079, 2007.

\bibitem{jean-bastien2020bootstrap}
Grill Jean-Bastien, Strub Florian, Altché Florent, Tallec Corentin,
  Pierre~Richemond H., Buchatskaya Elena, Doersch Carl, Bernardo~Pires Avila,
  Zhaohan~Guo Daniel, Mohammad~Azar Gheshlaghi, Piot Bilal, Kavukcuoglu Koray,
  Munos Rémi, and Valko Michal.
\newblock Bootstrap your own latent - a new approach to self-supervised
  learning.
\newblock {\em NIPS 2020}, 2020.

\bibitem{glenn_jocher_2020_4154370}
Glenn Jocher.
\newblock {ultralytics/yolov5: v3.1 - Bug Fixes and Performance Improvements}.
\newblock \url{https://github.com/ultralytics/yolov5}, Oct. 2020.

\bibitem{ke2004efficient}
Yan Ke, Rahul Sukthankar, and Larry Huston.
\newblock Efficient near-duplicate detection and sub-image retrieval.
\newblock In {\em ACM multimedia}, volume~4, page~5. Citeseer, 2004.

\bibitem{kim2003content}
Changick Kim.
\newblock Content-based image copy detection.
\newblock {\em Signal Processing: Image Communication}, 18(3):169--184, 2003.

\bibitem{kim2020proxy}
Sungyeon Kim, Dongwon Kim, Minsu Cho, and Suha Kwak.
\newblock Proxy anchor loss for deep metric learning.
\newblock In {\em Proceedings of the IEEE/CVF Conference on Computer Vision and
  Pattern Recognition}, pages 3238--3247, 2020.

\bibitem{DBLP:journals/corr/KingmaB14}
Diederik~P. Kingma and Jimmy Ba.
\newblock Adam: A method for stochastic optimization.
\newblock In {\em ICLR (Poster)}, 2015.

\bibitem{li2021momentum}
Zeming Li, Songtao Liu, and Jian Sun.
\newblock Momentum\^{} 2 teacher: Momentum teacher with momentum statistics for
  self-supervised learning.
\newblock {\em arXiv preprint arXiv:2101.07525}, 2021.

\bibitem{Liu_2017_CVPR}
Weiyang Liu, Yandong Wen, Zhiding Yu, Ming Li, Bhiksha Raj, and Le Song.
\newblock Sphereface: Deep hypersphere embedding for face recognition.
\newblock In {\em The IEEE Conference on Computer Vision and Pattern
  Recognition (CVPR)}, 2017.

\bibitem{liu2020content}
Xiaolong Liu, Jinchao Liang, Zi-Yi Wang, Yi-Te Tsai, Chia-Chen Lin, and
  Chih-Cheng Chen.
\newblock Content-based image copy detection using convolutional neural
  network.
\newblock {\em Electronics}, 9(12):2029, 2020.

\bibitem{oh2016deep}
Hyun Oh~Song, Yu Xiang, Stefanie Jegelka, and Silvio Savarese.
\newblock Deep metric learning via lifted structured feature embedding.
\newblock In {\em Proceedings of the IEEE conference on computer vision and
  pattern recognition}, pages 4004--4012, 2016.

\bibitem{oliva2001modeling}
Aude Oliva and Antonio Torralba.
\newblock Modeling the shape of the scene: A holistic representation of the
  spatial envelope.
\newblock {\em International journal of computer vision}, 42(3):145--175, 2001.

\bibitem{Pan_2018_ECCV}
Xingang Pan, Ping Luo, Jianping Shi, and Xiaoou Tang.
\newblock Two at once: Enhancing learning and generalization capacities via
  ibn-net.
\newblock In {\em Proceedings of the European Conference on Computer Vision
  (ECCV)}, September 2018.

\bibitem{qian2019softtriple}
Qi Qian, Lei Shang, Baigui Sun, Juhua Hu, Hao Li, and Rong Jin.
\newblock Softtriple loss: Deep metric learning without triplet sampling.
\newblock In {\em Proceedings of the IEEE/CVF International Conference on
  Computer Vision}, pages 6450--6458, 2019.

\bibitem{radenovic2018fine}
Filip Radenovi{\'c}, Giorgos Tolias, and Ond{\v{r}}ej Chum.
\newblock Fine-tuning cnn image retrieval with no human annotation.
\newblock {\em IEEE transactions on pattern analysis and machine intelligence},
  41(7):1655--1668, 2018.

\bibitem{sohn2016improved}
Kihyuk Sohn.
\newblock Improved deep metric learning with multi-class n-pair loss objective.
\newblock In {\em Advances in neural information processing systems}, pages
  1857--1865, 2016.

\bibitem{Sun_2020_CVPR}
Yifan Sun, Changmao Cheng, Yuhan Zhang, Chi Zhang, Liang Zheng, Zhongdao Wang,
  and Yichen Wei.
\newblock Circle loss: A unified perspective of pair similarity optimization.
\newblock In {\em Proceedings of the IEEE/CVF Conference on Computer Vision and
  Pattern Recognition (CVPR)}, June 2020.

\bibitem{sun2018beyond}
Yifan Sun, Liang Zheng, Yi Yang, Qi Tian, and Shengjin Wang.
\newblock Beyond part models: Person retrieval with refined part pooling (and a
  strong convolutional baseline).
\newblock In {\em Proceedings of the European conference on computer vision
  (ECCV)}, pages 480--496, 2018.

\bibitem{szegedy2016rethinking}
Christian Szegedy, Vincent Vanhoucke, Sergey Ioffe, Jon Shlens, and Zbigniew
  Wojna.
\newblock Rethinking the inception architecture for computer vision.
\newblock In {\em Proceedings of the IEEE conference on computer vision and
  pattern recognition}, pages 2818--2826, 2016.

\bibitem{uijlings2013selective}
Jasper~RR Uijlings, Koen~EA Van De~Sande, Theo Gevers, and Arnold~WM Smeulders.
\newblock Selective search for object recognition.
\newblock {\em International journal of computer vision}, 104(2):154--171,
  2013.

\bibitem{wan2008survey}
YH Wan, QL Yuan, SM Ji, LM He, and YL Wang.
\newblock A survey of the image copy detection.
\newblock In {\em 2008 IEEE Conference on Cybernetics and Intelligent Systems},
  pages 738--743. IEEE, 2008.

\bibitem{wang2017normface}
Feng Wang, Xiang Xiang, Jian Cheng, and Alan~Loddon Yuille.
\newblock Normface: L2 hypersphere embedding for face verification.
\newblock In {\em Proceedings of the 25th ACM international conference on
  Multimedia}, pages 1041--1049, 2017.

\bibitem{wang2018cosface}
Hao Wang, Yitong Wang, Zheng Zhou, Xing Ji, Dihong Gong, Jingchao Zhou, Zhifeng
  Li, and Wei Liu.
\newblock Cosface: Large margin cosine loss for deep face recognition.
\newblock In {\em Proceedings of the IEEE conference on computer vision and
  pattern recognition}, pages 5265--5274, 2018.

\bibitem{wang2021domainmix}
Wenhao Wang, Shengcai Liao, Fang Zhao, Kangkang Cui, and Ling Shao.
\newblock Domainmix: Learning generalizable person re-identification without
  human annotations.
\newblock In {\em British Machine Vision Conference}, 2021.

\bibitem{wang2020attentive}
Wenhao Wang, Fang Zhao, Shengcai Liao, and Ling Shao.
\newblock Attentive waveblock: Complementarity-enhanced mutual networks for
  unsupervised domain adaptation in person re-identification and beyond.
\newblock {\em arXiv preprint arXiv:2006.06525}, 2020.

\bibitem{Zbontar2021BarlowTS}
Jure Zbontar, Li Jing, Ishan Misra, Yann LeCun, and St{\'e}phane Deny.
\newblock Barlow twins: Self-supervised learning via redundancy reduction.
\newblock In {\em ICML}, 2021.

\bibitem{zheng2020vehiclenet}
Zhedong Zheng, Tao Ruan, Yunchao Wei, Yi Yang, and Tao Mei.
\newblock Vehiclenet: Learning robust visual representation for vehicle
  re-identification.
\newblock {\em IEEE Transactions on Multimedia}, 2020.

\bibitem{zhou2020cnn}
Zhili Zhou, Meimin Wang, Yi Cao, and Yuecheng Su.
\newblock Cnn feature-based image copy detection with contextual hash
  embedding.
\newblock {\em Mathematics}, 8(7):1172, 2020.

\end{thebibliography}
}

\end{document}